\documentclass{article}
\usepackage[preprint]{colm2025_conference}

\usepackage[utf8]{inputenc}
\usepackage[T1]{fontenc}
\usepackage{booktabs}
\usepackage{amsfonts}
\usepackage{amsmath}
\usepackage{graphicx}
\usepackage{subcaption}
\usepackage{xcolor}
\usepackage{multirow}
\usepackage[colorlinks=true,linkcolor=blue,citecolor=blue,urlcolor=blue]{hyperref}
\newcommand{\R}{\mathbb{R}}
\newcommand{\E}{\mathbb{E}}

\newcommand{\vecs}{\mathbf{v}_{\text{seal}}}
\newcommand{\vstable}{\mathbf{v}_{\text{stable}}}
\newcommand{\vproj}{\mathbf{v}_{\text{proj}}}
\newcommand{\vcombined}{\mathbf{v}_{\text{A+B}}}
\newcommand{\hstate}{\mathbf{h}}
\newcommand{\dv}{\mathbf{d}}

\title{Reliable Control-Point Selection for Steering Reasoning in Large Language Models}

\author{
  \begin{tabular}[t]{c}
  \bf Haomin Zhuang, Hojun Yoo, Xiaonan Luo, Kehan Guo, Xiangliang Zhang \\
  University of Notre Dame \\
  \texttt{\{hzhuang2, xzhang33\}@nd.edu}
  \end{tabular}
}

\begin{document}
\maketitle

\begin{abstract}
Steering vectors offer a training-free mechanism for controlling reasoning behaviors in large language models, but constructing effective vectors requires identifying genuine behavioral signals in the model's hidden states. For behaviors that can be toggled via prompts, this is straightforward. However, many reasoning behaviors---such as self-reflection---emerge spontaneously and resist prompt-level control. Current methods detect these behaviors through keyword matching in chain-of-thought traces, implicitly assuming that every detected boundary encodes a genuine behavioral signal. We show that this assumption is overwhelmingly wrong: across 541 keyword-detected boundaries, 93.3\% are behaviorally unstable, failing to reproduce the detected behavior under re-generation from the same prefix. We develop a probabilistic model that formalizes intrinsic reasoning behaviors as stochastic events with context-dependent trigger probabilities, and show that unstable boundaries dilute the steering signal. Guided by this analysis, we propose stability filtering, which retains only boundaries where the model consistently reproduces the target behavior. Combined with a content-subspace projection that removes residual question-specific noise, our method achieves 0.784 accuracy on MATH-500 (+5.0 over the strongest baseline). The resulting steering vectors transfer across models in the same architecture family without re-extraction, improving Nemotron-Research-Reasoning-1.5B (+5.0) and DeepScaleR-1.5B-Preview (+6.0). Code is available at \url{https://github.com/zhmzm/stability-steering}.
\end{abstract}

\section{Introduction}

Recent large language models solve complex tasks by generating extended chains of thought before producing a final answer~\citep{wei2022chain}. During this reasoning process, models spontaneously exhibit distinct behavioral patterns---reflecting on intermediate results, switching strategies when stuck, verifying conclusions before committing~\citep{chen2025seal}. These behaviors often improve accuracy, but they can also be counterproductive: models frequently over-reflect or repeat unproductive steps, wasting tokens without improving solutions~\citep{huang2025mitigating,sui2025overthinking}. Steering vectors offer a training-free mechanism for controlling such behaviors by adding learned directions to hidden states during generation~\citep{turner2023steering,zou2023representation}. The standard approach constructs a steering direction by contrasting hidden states with and without the target behavior, then applies this direction at inference time.

Constructing effective steering vectors in such manners requires identifying where genuine behavioral signals are encoded. For behaviors that can be toggled via prompts---such as output format or response style---this is straightforward: one contrasts hidden states with and without the target instruction. However, many reasoning behaviors resist prompt-level control. A model may pause to verify an intermediate result, yet instructing it to ``reflect less'' neither suppresses the pattern nor yields a useful steering direction (\S\ref{sec:main_results}). For such behaviors, one must instead extract signals from the model's own reasoning traces, typically by detecting behavioral boundaries through surface-level cues such as keyword matching~\citep{chen2025seal}. This raises a fundamental question: \emph{does a detected boundary actually reflect a genuine behavioral commitment by the model?}

We find that, overwhelmingly, it does not. The core issue is that intrinsic reasoning behaviors are stochastic: at any given position, a model may or may not produce the target behavior across independent generations. A keyword match at such a position is therefore a noisy proxy---it may capture a genuine behavioral commitment or a coincidental surface pattern. Through systematic re-generation experiments across 541 keyword-detected boundaries, we find that 93.3\% are behaviorally unstable: the detected behavior fails to reproduce when the model is re-run from the same prefix. The mean trigger probability is only $\bar{p} = 0.167$, indicating that existing methods retain roughly one-sixth of the maximum possible behavioral signal.

\begin{figure*}[t]
    \centering
    \includegraphics[width=\linewidth]{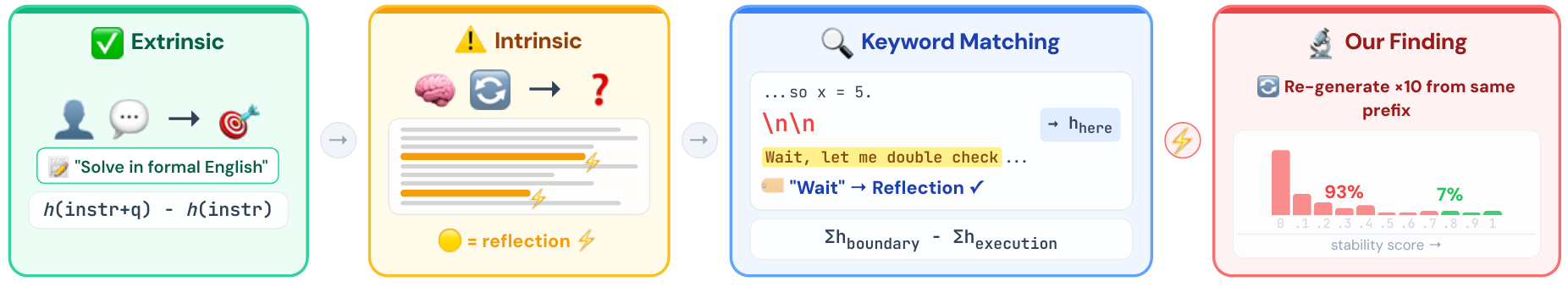}
    \caption{Extrinsic behaviors are steered via prompt contrast (left). Intrinsic behaviors emerge spontaneously and resist prompting (middle-left). Keyword matching detects them from traces (middle-right), but most detections are unstable under re-generation (right; data from \S\ref{sec:stability_results}).}
    \label{fig:concept}
\end{figure*}

This observation motivates a simple principle: \emph{only behaviorally stable boundaries should contribute to the steering vector}. We estimate each boundary's trigger probability by re-generating from its prefix multiple times, then retain only those that consistently reproduce the target behavior. A probabilistic analysis shows that this filtering amplifies the behavioral signal in proportion to the ratio of filtered to unfiltered trigger probabilities, and predicts that hard thresholding should outperform soft weighting when the trigger distribution is skewed---both predictions are confirmed experimentally. Combined with a content-subspace projection that removes residual question-specific noise, our method yields substantial improvements over the strongest existing baseline on MATH-500, and the resulting steering vectors can be transferred across models within the same architecture family without re-extraction
.

Our contributions are summarized as follows:
\begin{enumerate}
    \item We develop a \textbf{probabilistic model of intrinsic reasoning behaviors} that formalizes them as stochastic events with context-dependent trigger probabilities, explains why surface-detected boundaries dilute the steering signal, and yields testable predictions about filtering strategies.

    \item We propose \textbf{stability filtering}, a method that estimates trigger probabilities via repeated sampling and retains only high-probability boundaries. Combined with content-subspace projection, this achieves state-of-the-art accuracy on MATH-500 across three models. A behavior probe confirms that stable boundaries carry stronger behavioral signals in their hidden states.

    \item We demonstrate \textbf{cross-model transfer}: steering vectors extracted once from a single source model improve two additional models in the same architecture family, without any model-specific re-extraction.
\end{enumerate}

\section{Related Work}
\vspace{-0.01in}
\paragraph{Reasoning in LLMs.}
Chain-of-thought prompting~\citep{wei2022chain} established that LLMs can solve complex reasoning tasks by generating intermediate steps, and subsequent work has extended this through self-consistency~\citep{wang2023selfconsistency}, tree-of-thought search~\citep{yao2023tree}, step-level verification~\citep{lightman2023lets}, and self-reflection~\citep{huang2023large,ma2025s2r}. Reinforcement learning has further amplified these capabilities: DeepSeek-R1~\citep{guo2025deepseek} uses RL without supervised fine-tuning to induce self-verification and extended chain-of-thought, achieving strong performance on math and code benchmarks. Test-time compute scaling~\citep{zeng2025revisiting} shows that allocating more inference tokens improves reasoning, but this comes at a cost. Recent work has identified an \emph{overthinking} problem~\citep{sui2025overthinking,huang2025mitigating}: models frequently over-reflect or repeat unproductive reasoning steps, wasting tokens without improving accuracy. Controlling these intrinsic reasoning behaviors is the central motivation of our work.

\vspace{-0.01in}
\paragraph{Steering vectors for LLMs.}
Controllable generation in LLMs has been explored through both training-time and inference-time interventions~\citep{dathathri2019plug}. Activation steering~\citep{turner2023steering} showed that adding fixed directions to residual stream activations can induce behavioral changes, and subsequent work has applied this principle to truthfulness~\citep{li2023iti}, safety and refusal~\citep{arditi2024refusal}, personality traits~\citep{chen2025persona}, and contrastive behavior control~\citep{rimsky2024caa}. \citet{zou2023representation} proposed representation engineering as a general framework for monitoring and manipulating high-level cognitive phenomena in LLMs. The theoretical basis for these methods rests on the linear representation hypothesis~\citep{park2024linear,burns2022discovering}, which posits that high-level concepts are encoded as linear directions in residual stream space. For reasoning specifically, SEAL~\citep{chen2025seal} introduced a lightweight approach to calibrating reasoning behaviors via hidden-state contrasts extracted from chain-of-thought traces, and \citet{sinii2025small} provided a mechanistic analysis of how steering vectors modify reasoning. \citet{venhoff2025understanding} systematically mapped steering directions to distinct reasoning behaviors. Our work builds directly on SEAL's pipeline but addresses a fundamental limitation: the assumption that keyword-detected boundaries all encode genuine behavioral signals.

\vspace{-0.01in}

\paragraph{Stochastic behaviors and robustness.}
Our work is grounded in the observation that intrinsic reasoning behaviors are stochastic rather than deterministic. \citet{dies2025stability} showed that LLM belief representations can be unstable under semantic perturbations, and \citet{wang2023interpretability} noted that behavioral features in language models vary across sampling conditions. \citet{perez2022discovering} demonstrated that model-written evaluations can reveal inconsistent behaviors that are not apparent from single-sample analysis. Our stability probing methodology is related to consistency-based filtering in data curation~\citep{swayamdipta2020dataset}, where training examples are classified by their learning dynamics. We apply an analogous principle to steering vector construction: only boundaries where the model stochastically reproduces the target behavior with high probability should contribute to the steering direction.

\section{Method}

\begin{figure*}[t]
    \centering
    \includegraphics[width=0.90\linewidth]{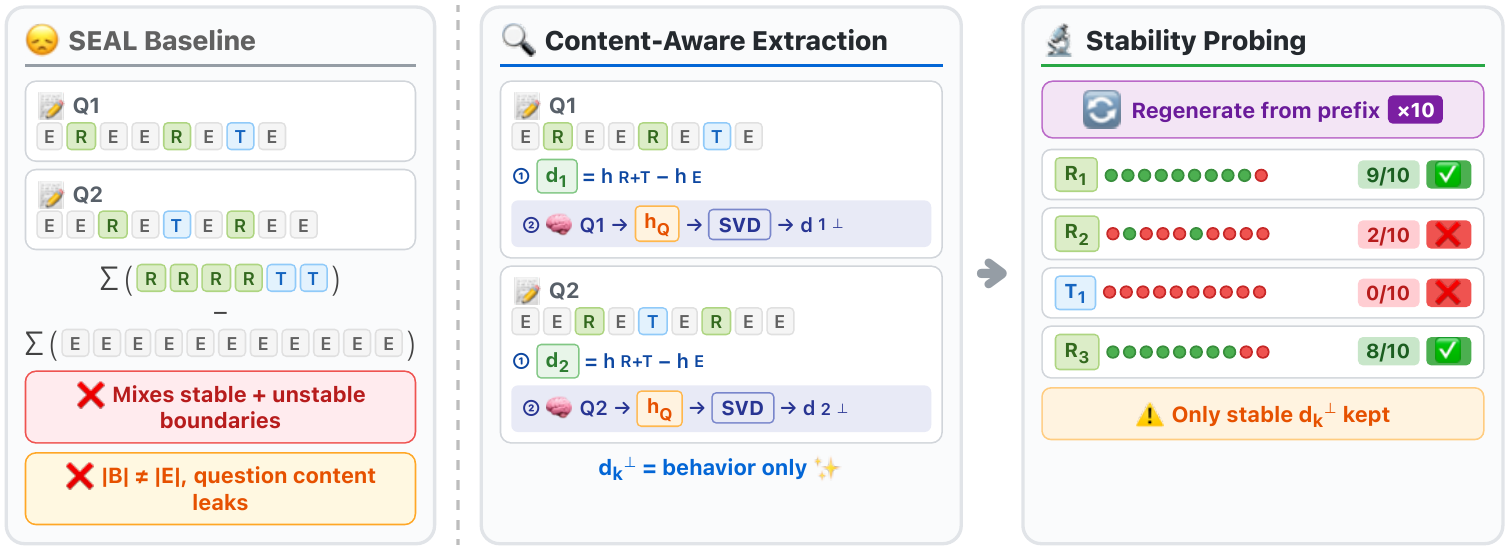}
    \caption{Overview of our method. \textbf{Left}: SEAL mixes all keyword-detected boundaries, including unstable ones that dilute the signal. \textbf{Middle}: Our content-aware extraction computes per-question steering directions and projects out question-specific information via SVD. \textbf{Right}: Stability probing re-generates from each boundary's prefix and retains only those that reliably reproduce the target behavior.}
    \label{fig:pipeline}
\end{figure*}

Figure~\ref{fig:pipeline} gives an overview of our approach. We start from the SEAL pipeline (\S\ref{sec:prelim}), identify why it underperforms through a probabilistic lens (\S\ref{sec:theory}), and propose two fixes: stability filtering (\S\ref{sec:robustness}) and content projection (\S\ref{sec:content}).

\subsection{Preliminaries: SEAL Steering Vectors}
\label{sec:prelim}

Given a reasoning LLM and a problem set $\mathcal{Q} = \{q_1, \ldots, q_N\}$, SEAL~\citep{chen2025seal} constructs a steering vector as follows. For each problem $q_k$, the model generates a chain-of-thought response, which is segmented into paragraphs. Each paragraph is classified as a \emph{behavior boundary} or an \emph{execution} segment based on the presence of reasoning-related keywords (e.g., ``wait,'' ``verify,'' ``alternatively''), yielding boundary positions $B_k$ and execution positions $E_k$. All hidden states below are extracted at the first token of each paragraph at a target layer $\ell$; we write $\hstate_b \in \R^D$ and suppress the layer index for clarity.

A per-example steering vector is computed as:
\begin{equation}
    \dv_k = \frac{1}{|B_k|}\sum_{b \in B_k} \hstate_b - \frac{1}{|E_k|}\sum_{e \in E_k} \hstate_e,
    \label{eq:seal}
\end{equation}
where the first term averages over behavior boundary hidden states and the second over execution hidden states. The final SEAL vector is $\vecs = \text{normalize}(\frac{1}{N}\sum_k \dv_k)$. Problems where $|B_k| = 0$ or $|E_k| = 0$ are excluded. At inference, the steering vector is applied at paragraph boundaries within the \texttt{<think>} block: $\hstate \leftarrow \hstate + \alpha \cdot \vecs$, using the same keyword rules for boundary detection.

\subsection{Why SEAL's Signal Is Diluted}
\label{sec:theory}

The core issue with SEAL is that keyword matching is a poor proxy for genuine behavior. A paragraph containing ``wait'' may or may not reflect an actual reflective reasoning step. We formalize this with a simple probabilistic model. The analysis below uses idealizing assumptions and is meant as motivation, not as a formal guarantee, we test its predictions experimentally in \S\ref{sec:stability_results}--\ref{sec:filtering}.

\paragraph{Setup.}
Recall from \S1 that \emph{intrinsic behaviors} emerge spontaneously during reasoning and cannot be reliably controlled through prompts. At each candidate boundary $b$ with context $c_b$, we define a trigger probability $p_b = P(\text{behavior } a \mid c_b)$, where $a$ is the event that the model produces a genuine behavior boundary. Following SEAL, we merge Reflection (R) and Transition (T) into one class and treat the problem as binary. Since $p_b$ is not directly observable, keyword detection is a noisy proxy; our stability probing (\S\ref{sec:robustness}) estimates $p_b$ empirically.

We model the hidden state at boundary $b$ with a latent commitment indicator $z_b \sim \text{Bernoulli}(p_b)$:
\begin{equation}
    \hstate_b = z_b \cdot \boldsymbol{\delta} + \boldsymbol{\mu}(c_b) + \boldsymbol{\epsilon}_b,
    \label{eq:decomp}
\end{equation}
where $\boldsymbol{\delta} \in \R^D$ is a shared behavioral direction, $\boldsymbol{\mu}(c_b)$ is context-dependent (question content, reasoning state), and $\boldsymbol{\epsilon}_b$ is residual noise. Execution positions have no behavioral signal: $\hstate_e = \boldsymbol{\mu}(c_e) + \boldsymbol{\epsilon}_e$. The upshot is that the behavioral component $\boldsymbol{\delta}$ appears in $\hstate_b$ only to the extent that the model is actually committed to the behavior at that point, which happens with probability $p_b$.

\paragraph{Dilution.}
We make three simplifying assumptions:
\begin{enumerate}
    \item[(A1)] Context terms roughly cancel between boundary and execution means within each problem: $\frac{1}{|B_k|}\sum_{b} \boldsymbol{\mu}(c_b) \approx \frac{1}{|E_k|}\sum_{e} \boldsymbol{\mu}(c_e)$. This is a strong assumption; we revisit it in \S\ref{sec:content}.
    \item[(A2)] Noise terms $\boldsymbol{\epsilon}_b$ are approximately independent with zero mean.
    \item[(A3)] Execution positions carry no behavioral signal ($z_e \equiv 0$).
\end{enumerate}
Substituting into Eq.~\ref{eq:seal}:
\begin{equation}
    \E[\dv_k] = \frac{1}{|B_k|}\sum_{b} p_b \cdot \boldsymbol{\delta} + \underbrace{\left(\frac{1}{|B_k|}\sum_b \boldsymbol{\mu}(c_b) - \frac{1}{|E_k|}\sum_e \boldsymbol{\mu}(c_e)\right)}_{\approx\;\mathbf{0}\text{ by (A1)}} \;\approx\; \bar{p}_k \cdot \boldsymbol{\delta},
    \label{eq:dilution}
\end{equation}
where $\bar{p}_k = \frac{1}{|B_k|}\sum_{b \in B_k} p_b$. The SEAL vector $\vecs$ is normalized, so the scalar $\bar{p}$ disappears from the final direction. But that does not make the problem go away: when $\bar{p}$ is small, the noise terms $\boldsymbol{\epsilon}_b$ dominate the signal \emph{before} normalization, so the resulting direction is poorly aligned with $\boldsymbol{\delta}$.

\paragraph{Filtering as a fix.}
If we restrict to boundaries with $p_b \geq \tau$, writing $B_k^{\tau} = \{b \in B_k : p_b \geq \tau\}$:
\begin{equation}
    \E[\dv_k^{\text{stable}}] \approx \bar{p}_{k,\tau} \cdot \boldsymbol{\delta}, \quad \bar{p}_{k,\tau} = \frac{1}{|B_k^{\tau}|}\sum_{b \in B_k^{\tau}} p_b.
    \label{eq:amplification}
\end{equation}
The signal-to-noise ratio before normalization improves by roughly $\bar{p}_\tau / \bar{p}$ (assuming comparable per-boundary noise). When $\bar{p}$ is small and $\bar{p}_\tau \approx 1$, this can be a large factor, leading to much better directional alignment with $\boldsymbol{\delta}$ after normalization. We test this in \S\ref{sec:filtering}.

One might instead use \emph{soft weighting}, letting each boundary contribute proportionally to $p_b$. Under (A1)--(A3) the soft-weighted signal coefficient works out to $\E[p] + \text{Var}(p)/\E[p]$, where expectations are over the empirical distribution of $p_b$ across all $|B| = \sum_k |B_k|$ boundaries. This helps somewhat, but still lets low-$p_b$ boundaries inject noise. Hard thresholding at $\tau$ guarantees a signal coefficient of at least $\tau$, at the cost of reducing available data from $|B|$ to $|B^{\tau}| = \sum_k |B_k^{\tau}|$. When the $p_b$ distribution is highly skewed, the bias reduction outweighs the variance cost, provided enough boundaries survive. Our experiments confirm this: hard filtering at $\tau{=}0.8$ substantially outperforms soft weighting (\S\ref{sec:stability_results}).

\subsection{Robustness-Guided Boundary Filtering}
\label{sec:robustness}

The analysis above needs $p_b$, which we cannot observe directly. We estimate it with a simple re-generation procedure.

For each candidate boundary $b$, we take the prefix up to $b$ and sample $M = 10$ continuations (temperature 0.7, top-$p$ 0.95, max 128 new tokens). The stability score is the fraction of continuations that reproduce a behavior boundary:
\begin{equation}
    s(b) = \hat{p}_b = \frac{1}{M} \sum_{i=1}^{M} \mathbb{1}[\text{behavior boundary in continuation } i].
    \label{eq:stability}
\end{equation}
Ten samples is admittedly coarse: the standard error is at most 0.16, giving a 95\% CI half-width of ${\sim}0.31$. But this is enough to tell apart $p_b \approx 0$ from $p_b \geq 0.8$, which is the distinction that matters most. The threshold $\tau$ should be understood as approximate.

We keep only boundaries with $s(b) \geq \tau$ and discard the rest:
\begin{equation}
    \dv_k^{\text{stable}} = \frac{1}{|B_k^{\tau}|}\sum_{b \in B_k^{\tau}} \hstate_b - \frac{1}{|E_k|}\sum_{e \in E_k} \hstate_e,
    \label{eq:stable}
\end{equation}
where $B_k^{\tau} = \{b \in B_k : s(b) \geq \tau\}$. Problems with no surviving boundaries are dropped. The filtered vector is $\vstable = \text{normalize}(\frac{1}{N'}\sum_k \dv_k^{\text{stable}})$, where $N' \leq N$ is the number of retained problems.

\subsection{Content-Subspace Projection}
\label{sec:content}

A separate source of noise in the steering vector is question-specific content. The SEAL per-example vector (Eq.~\ref{eq:seal}) subtracts the execution mean from the boundary mean, but the two groups are not aligned: they differ in size ($|B_k| \neq |E_k|$) and occupy systematically different positions in the reasoning chain. Boundaries tend to appear at harder or more uncertain junctures, while execution segments dominate routine computation. As a result, the subtraction does not fully cancel the context term $\boldsymbol{\mu}$, and the residual carries question-level information such as math subject or problem difficulty. To remove this, we estimate the ``question-specific directions'' from question-only hidden states and project them out. This is a heuristic: we do not know that the residual in $\dv_k$ falls exactly in this subspace, but question-only representations provide a reasonable proxy for the directions we want to suppress.

Concretely, for each problem $k$ we feed only the question text through the model and mean-pool hidden states at layer $\ell$ to get $\mathbf{q}_k \in \R^D$. We center the matrix $\mathbf{Q} \in \R^{N \times D}$ and take its SVD: $\mathbf{Q}_c = \mathbf{U}\boldsymbol{\Sigma}\mathbf{V}^\top$. The top-$K$ right singular vectors $\mathbf{V}_K \in \R^{D \times K}$ define the content subspace.

We assume that $\boldsymbol{\delta}$ has negligible projection onto this subspace. This is an approximation: if $\boldsymbol{\delta}$ does overlap, projection will attenuate the signal. The rank $K$ controls the trade-off; we use $K{=}4$ based on probing results (\S\ref{sec:probe}). Each steering vector is projected out:
\begin{equation}
    \dv_k^{\perp} = \dv_k - \mathbf{V}_K (\mathbf{V}_K^\top \dv_k),
    \label{eq:proj}
\end{equation}
and the projected vector is $\vproj = \text{normalize}(\frac{1}{N}\sum_k \dv_k^{\perp})$. By linearity, this is equivalent to projecting each $\hstate_b$ and $\hstate_e$ individually before computing the difference.

\subsection{The Final Steering Vector}
\label{sec:combine}

The final steering vector combines both fixes. We first filter for stable boundaries, then project out the content subspace:
\begin{equation}
    \vcombined = \text{normalize}\!\left(\frac{1}{N'}\sum_{k} \left[ \frac{1}{|B_k^{\tau}|}\sum_{b \in B_k^{\tau}} \hstate_b^{\perp} - \frac{1}{|E_k|}\sum_{e \in E_k} \hstate_e^{\perp} \right]\right),
    \label{eq:combined}
\end{equation}
where $\hstate^{\perp} = \hstate - \mathbf{V}_K(\mathbf{V}_K^\top \hstate)$ and $B_k^{\tau} = \{b \in B_k : s(b) \geq \tau\}$. Filtering improves the signal (better alignment with $\boldsymbol{\delta}$); projection aims to reduce residual content noise. We ablate their individual and combined contributions in \S\ref{sec:main_results}.

\section{Experiments}

\subsection{Setup}

\paragraph{Models.} We extract steering vectors from DeepSeek-R1-Distill-Qwen-1.5B and evaluate on the same model. To test cross-model transfer, we apply the same vectors (without re-extraction) to two additional Qwen-architecture 1.5B models: Nemotron-Research-Reasoning-1.5B and DeepScaleR-1.5B-Preview. All models produce chain-of-thought reasoning within \texttt{<think>} blocks.

\paragraph{Data and evaluation protocol.} Steering vectors are extracted from 100 MATH training problems and evaluated on MATH-500 (test set, 500 questions). All main results (Table~\ref{tab:main_results}) use \texttt{max\_tokens=4096} with greedy decoding (temperature 0). Accuracy is measured by exact match against the ground-truth answer after extracting the final \texttt{\textbackslash boxed\{\}} expression.

\paragraph{Hyperparameters.} Steering is applied at layer 20, corresponding to 71\% depth in the 28-layer Qwen architecture. The steering coefficient $\alpha{=}-100$ is selected on a held-out set. The stability threshold $\tau{=}0.8$ is selected from a sweep at \texttt{max\_tokens=1024} (\S\ref{sec:filtering}) and then held fixed for the main 4096-token evaluation; the content-subspace rank $K{=}4$ is chosen based on the probing analysis in \S\ref{sec:probe}. Full coefficient sweeps for all methods are reported in Appendix~\ref{app:coef_sweep} and~\ref{app:stable_coef}.

\subsection{Main Results}
\label{sec:main_results}

\begin{table}[t]
\centering
\caption{Main results on MATH-500 (500 questions, max\_tokens=4096, layer 20). Steering vectors extracted from DeepSeek-R1-Distill-Qwen-1.5B. $\Delta$ = accuracy change vs.\ each model's own baseline. Avg Tok = average word count; Avg R = average reflection keyword count.}
\label{tab:main_results}
\setlength{\tabcolsep}{4pt}
\small
\begin{tabular}{llcccc}
\toprule
\textbf{Model} & \textbf{Method} & \textbf{Acc} & \textbf{$\Delta$} & \textbf{Avg Tok} & \textbf{Avg R} \\
\midrule
\multicolumn{6}{l}{\emph{Source model (vectors extracted here)}} \\
\multirow{5}{*}{Qwen-1.5B}
  & Baseline      & 0.608 & ---              & 1547 & 39.1 \\
  & Prompt text   & 0.672 & +6.4             & 1116 & 24.3 \\
  & Prompt vector & 0.610 & +0.2             & 1604 & 39.4 \\
  & SEAL          & 0.734 & +12.6            & 1103 & 10.3 \\
  & Ours          & \textbf{0.784} & \textbf{+17.6} & 1008 & 10.7 \\
\midrule
\multicolumn{6}{l}{\emph{Cross-model transfer (same vectors, no re-extraction)}} \\
\multirow{3}{*}{Nemotron-1.5B}
  & Baseline & 0.662 & ---              & 1660 & 76.9 \\
  & SEAL     & 0.766 & +10.4            & 1230 & 43.7 \\
  & Ours     & \textbf{0.816} & \textbf{+15.4} & 1147 & 39.8 \\
\multirow{3}{*}{DeepScaleR-1.5B}
  & Baseline & 0.726 & ---              & 1323 & 35.1 \\
  & SEAL     & 0.752 & +2.6             & 1076 & 17.9 \\
  & Ours     & \textbf{0.812} & \textbf{+8.6}  & 1016 & 14.5 \\
\bottomrule
\end{tabular}
\vspace{-0.1in}
\end{table}

Table~\ref{tab:main_results} presents our main results on MATH-500 with \texttt{max\_tokens=4096}. The unsteered baseline achieves 0.608. SEAL improves this to 0.734 (+12.6). Our method reaches \textbf{0.784} (+17.6 over baseline), a further +5.0 beyond SEAL. Compared to the baseline, steering reduces average word count from 1547 to 1008 and average reflection keywords from 39.1 to 10.7, indicating less unproductive reasoning. Our method maintains a similar level of reflection suppression as SEAL (Avg R = 10.7 vs.\ 10.3); the accuracy gain comes primarily from stability filtering, as shown by the component ablation in Table~\ref{tab:ablation}. Stability filtering (B) accounts for +7.0 over SEAL, while content projection alone (A) yields a modest +1.6. Combining both gives +7.8, confirming that filtering is the primary driver.

\paragraph{Prompt baselines.}
To ground the intrinsic/extrinsic distinction from \S1, we include two prompt-based interventions in the same table. Adding ``use less reflection'' to the prompt yields 0.672 (+6.4), and a prompt-derived steering vector ($\mathbf{h}_{Q+\text{prompt}} - \mathbf{h}_Q$) achieves only 0.610 (+0.2). Both fall well below activation steering (0.784). While we tested only a limited set of prompt formulations, these results are consistent with reflection being an intrinsic behavior that resists prompt-level control.

\paragraph{Cross-model transfer.}
The same steering vectors, extracted once from Qwen-1.5B, also work on other Qwen-architecture 1.5B models without re-extraction. On Nemotron-Research-Reasoning-1.5B, our method achieves 0.816 (+15.4 over baseline) vs.\ SEAL's 0.766. On DeepScaleR-1.5B-Preview, our method achieves 0.812 (+8.6) vs.\ SEAL's 0.752. These models share the same architecture and were trained on similar reasoning data, which likely explains why the steering direction transfers. Whether this extends to more distant model families remains an open question.

\subsection{How Unstable Are Keyword-Detected Boundaries?}
\label{sec:stability_results}

\begin{figure*}[t]
    \centering
    \begin{minipage}[t]{0.48\linewidth}
        \centering
        \includegraphics[width=\linewidth]{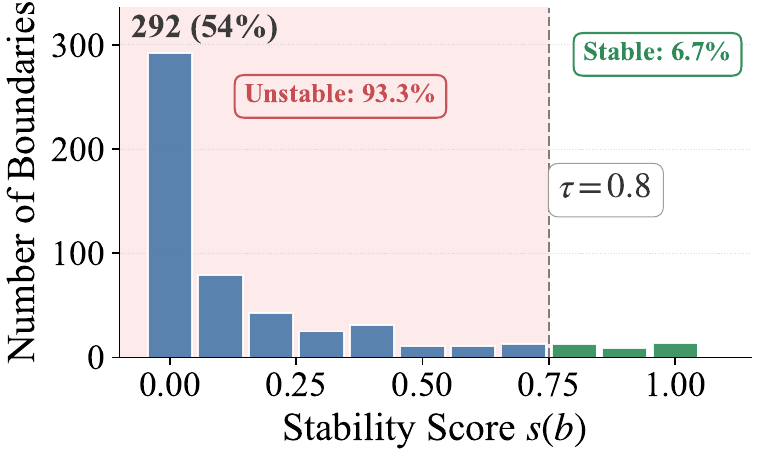}
        \centerline{(a) Stability score distribution}
    \end{minipage}
    \hfill
    \begin{minipage}[t]{0.48\linewidth}
        \centering
        \includegraphics[width=\linewidth]{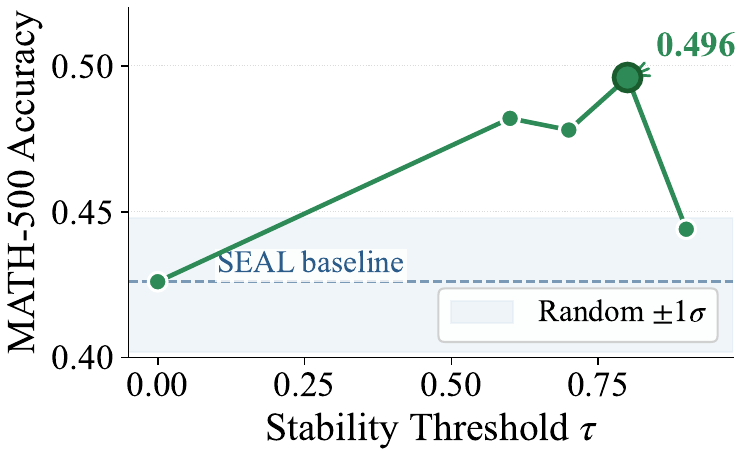}
        \centerline{(b) Accuracy vs.\ stability threshold $\tau$}
    \end{minipage}
    \caption{(a) Distribution of stability scores across 541 R+T boundaries. 93.3\% are unstable ($s(b) < 0.8$); only 6.7\% pass the threshold. (b) Accuracy peaks at $\tau{=}0.8$ (+7.0 over SEAL), then drops at $\tau{=}0.9$ as sample size shrinks. Shaded band shows random baseline $\pm 1\sigma$.}
    \label{fig:stability_distribution}
    \vspace{-0.1in}
\end{figure*}

If the stochastic model in \S\ref{sec:theory} is roughly right, we should expect two things: (i) most keyword-detected boundaries have low trigger probability $p_b$, and (ii) the distribution of $p_b$ should be heavily skewed rather than smoothly spread. Both predictions are borne out by Figure~\ref{fig:stability_distribution}(a).

Across 541 keyword-detected behavior boundaries (Reflection and Transition) extracted from 100 problems, 292 (54\%) have $s(b) = 0$: the behavior never reappears across 10 independent re-generations from the same prefix. Only 36 (6.7\%) meet or exceed $\tau{=}0.8$. The distribution is heavily zero-inflated with a long right tail: beyond the spike at zero, scores decay gradually through the 0.1--0.7 range before a small cluster at $s \geq 0.8$. This is consistent with keyword matching picking up a large population of coincidental matches ($p_b \approx 0$) alongside a small minority of genuine behavioral commitments.

The mean trigger probability is $\bar{p} = 0.167$. Substituting into our dilution analysis (Eq.~\ref{eq:dilution}), this implies the SEAL vector retains roughly 17\% of the maximum behavioral signal. The heavy concentration of mass near zero also explains why hard thresholding (+7.0 over SEAL) substantially outperforms soft weighting (+2.2): low-$p_b$ boundaries dominate the average under any continuous weighting scheme, and only a hard cutoff can exclude them entirely.

\subsection{Does Filtering Actually Help?}
\label{sec:filtering}

The theory predicts that filtering for high-$p_b$ boundaries should amplify the steering signal. We test this with a threshold sweep, then check two potential confounds.

\paragraph{Threshold sweep.}
Figure~\ref{fig:stability_distribution}(b) shows accuracy as a function of $\tau$, evaluated at \texttt{max\_tokens=1024} on MATH-500. We use a shorter token budget here for computational efficiency; the main results in Table~\ref{tab:main_results} use 4096. Performance rises steadily from $\tau{=}0$ (SEAL, 0.426) to $\tau{=}0.8$ (0.496, +7.0), then falls at $\tau{=}0.9$ as too few boundaries survive the filter. At $\tau{=}0.8$, 36 boundaries remain with $\bar{p}_{0.8} \approx 0.9$, giving a theoretical signal amplification of $0.9/0.167 \approx 5.4\times$. The rise-then-fall pattern matches the bias--variance trade-off predicted in \S\ref{sec:theory}: increasing $\tau$ reduces noise but eventually starves the estimator of data.

\paragraph{Random matched controls.}
One might worry that using fewer extraction boundaries somehow helps by itself. To check, we build five control vectors by randomly sampling the same number of boundaries. These controls average $0.425 \pm 0.023$, essentially identical to SEAL (0.426). The stability-filtered vector (0.496) lies $3.1\sigma$ above this baseline, so the gain comes from boundary quality, not quantity.

\begin{figure}[t]
    \centering
    \includegraphics[width=0.48\textwidth]{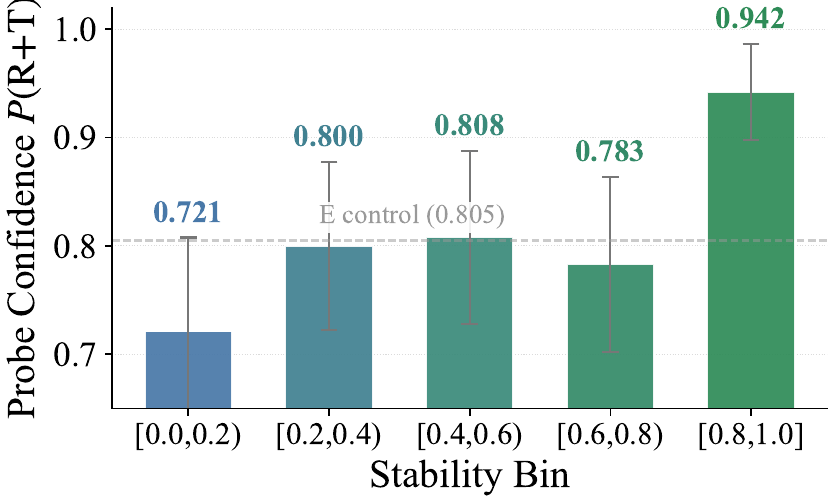}
    \caption{Behavior probe confidence by stability bin (balanced R+T vs.\ E, GroupKFold by question). Stable boundaries ($s \geq 0.8$) yield the highest confidence (0.942).}
    \label{fig:probe_confidence}
\end{figure}

\paragraph{Why does filtering help? A probing analysis.}
The accuracy gains above are consistent with our theory, but do stable boundaries actually carry a stronger behavioral signal in their hidden states? We train a balanced logistic regression probe to classify hidden states at layer 20 as behavior boundary vs.\ execution (methodology details in Appendix~\ref{app:probe_methodology}). Figure~\ref{fig:probe_confidence} shows that the most stable boundaries ($s \geq 0.8$) achieve a probe confidence of \textbf{0.942}, while the least stable ($s < 0.2$) score only 0.721 (overall accuracy 0.818, $\rho = {+}0.165$). This is what we expect from Eq.~\ref{eq:decomp}: higher $p_b$ means more $\boldsymbol{\delta}$ in the hidden state, which the probe picks up as a stronger boundary signal.

\subsection{Is the Content Subspace Separable from Behavior?}
\label{sec:probe}

Our content projection (\S\ref{sec:content}) assumes that question-specific information occupies a subspace approximately orthogonal to the behavioral direction $\boldsymbol{\delta}$. If this assumption fails, projection would attenuate the steering signal rather than clean it. We test this by probing the SVD-derived content subspace directly.

We decompose the 100 question hidden states under varying SVD ranks $k$ and train subject-classification probes (7 MATH categories) on the content component $Q_\parallel$ and residual $Q_\perp$ separately. Table~\ref{tab:svd_probe} shows a crossover. At low rank ($k \leq 2$), the content component performs worse than the residual, indicating the first few singular vectors capture noise rather than subject-discriminative content. From $k{=}4$ onward, the content component dominates: +14 at $k{=}4$, growing to +48 at $k{=}16$. The full hidden state achieves $Q_\text{full} = 0.650$ (vs.\ 7-class chance $\approx 0.143$). This supports our choice of $K{=}4$: the SVD subspace at this rank captures meaningful content information. We note that this probe validates the content subspace itself, not the stronger claim that steering vector contamination aligns with this subspace. The ablation in Table~\ref{tab:ablation} provides indirect evidence: combining projection with filtering yields a small additional gain (+0.8 over filtering alone), consistent with projection removing some residual noise, though the effect is modest.

\begin{table*}[t]
\centering
\begin{minipage}[t]{0.48\linewidth}
\centering
\caption{SVD content subspace probe (100 questions, 7 subjects, StratifiedKFold $n{=}5$). $Q_\text{full} = 0.650$; chance $\approx 0.143$.}
\label{tab:svd_probe}
\small
\begin{tabular}{rccc}
\toprule
$k$ & $Q_\parallel$ (content) & $Q_\perp$ (residual) & Sep. \\
\midrule
1  & 0.250 & 0.620 & $-37.0$ \\
2  & 0.320 & 0.600 & $-28.0$ \\
4  & 0.390 & 0.250 & $+14.0$ \\
8  & 0.500 & 0.110 & $+39.0$ \\
16 & 0.530 & 0.050 & $+48.0$ \\
\bottomrule
\end{tabular}
\end{minipage}
\hfill
\begin{minipage}[t]{0.48\linewidth}
\centering
\caption{Component ablation on MATH-500 (\texttt{max\_tokens=1024}, 500 questions). $\Delta$ vs.\ SEAL.}
\label{tab:ablation}
\small
\begin{tabular}{lcc}
\toprule
\textbf{Method} & \textbf{Acc} & \textbf{$\Delta$} \\
\midrule
Baseline (no steering) & 0.182 & --- \\
SEAL & 0.426 & --- \\
Content proj.\ only (A) & 0.442 & +1.6 \\
Stability filter only (B) & \textbf{0.496} & +7.0 \\
A+B (Ours) & \textbf{0.504} & +7.8 \\
\bottomrule
\end{tabular}
\end{minipage}
\end{table*}

\section{Conclusion}

We have shown that the majority of reasoning boundaries identified by keyword matching in LLM chain-of-thought traces are behaviorally unstable. 93.3\% fail to consistently reproduce the target reasoning behavior across re-generations. By filtering for behavioral stability and projecting out question-specific content, we construct steering vectors that achieve 0.784 accuracy on MATH-500, a 5\% improvement over SEAL. These vectors transfer effectively across models within the same architecture family, improving Nemotron-Research-Reasoning-1.5B by 5\% and DeepScaleR-1.5B-Preview by 6\% over their respective SEAL baselines. A balanced behavior probe confirms that stable boundaries encode cleaner behavioral signals in hidden states.

\bibliographystyle{colm2025_conference}
\bibliography{references}

\appendix
\section{Full Coefficient Sweep}
\label{app:coef_sweep}

Table~\ref{tab:full_coef} reports accuracy for SEAL and content-removed vectors across all steering coefficients on Qwen-1.5B.

\begin{table}[h]
\centering
\caption{Full coefficient sweep on MATH-500 (Qwen-1.5B, layer 20, \texttt{max\_tokens=1024}, 500 questions).}
\label{tab:full_coef}
\begin{tabular}{rcc}
\toprule
$\alpha$ & SEAL & Content Removal ($K{=}4$) \\
\midrule
$-100$ & 0.426 & 0.442 \\
$-50$ & 0.348 & 0.342 \\
$-20$ & 0.258 & 0.256 \\
$-1$ & 0.192 & 0.192 \\
Baseline & 0.182 & 0.182 \\
$1$ & 0.190 & 0.190 \\
$20$ & 0.126 & 0.122 \\
$50$ & 0.070 & 0.064 \\
$70$ & 0.038 & 0.046 \\
\bottomrule
\end{tabular}
\end{table}

\section{Coefficient Sweep for Stability-Filtered Vectors}
\label{app:stable_coef}

\begin{table}[h]
\centering
\caption{Coefficient sweep for $\vstable$ ($\tau{=}0.8$) and A+B combined on Qwen-1.5B (MATH-500, \texttt{max\_tokens=1024}, 500 questions).}
\begin{tabular}{rcc}
\toprule
$\alpha$ & $\vstable$ ($\tau{=}0.8$) & A+B Combined \\
\midrule
$-100$ & 0.496 & 0.504 \\
$-50$ & 0.364 & 0.366 \\
$-20$ & 0.254 & 0.260 \\
\bottomrule
\end{tabular}
\end{table}

The optimal coefficient is $\alpha{=}-100$ for both variants, consistent with the SEAL baseline.

\section{Probe Methodology Details}
\label{app:probe_methodology}

\paragraph{Behavior probe (R+T vs.\ E).}
We train a logistic regression classifier (\texttt{LogisticRegression(max\_iter=2000, C=1.0, solver=`lbfgs')}) to distinguish behavior boundary (R+T) hidden states from execution (E) hidden states at layer 20. We partition the 541 R+T boundaries into 5 equal stability bins of width 0.2 on $[0, 1)$ and sample 22 boundaries per bin, yielding 110 R+T samples total. We then downsample E to 110 samples, giving a balanced dataset of 220 total samples spanning 88 unique questions. Cross-validation uses GroupKFold ($n{=}5$ folds) grouped by \texttt{question\_id} to prevent data leakage across folds.

\paragraph{SVD content subspace probe.}
We probe 100 question hidden states (one per extraction question) decomposed under varying truncation ranks $k$. The 100 questions span 7 MATH subjects; we use StratifiedKFold ($n{=}5$) by subject label. For each $k$, we train separate probes on the rank-$k$ projection $Q_\parallel$ and the orthogonal residual $Q_\perp$, measuring 7-class subject classification accuracy.

\end{document}